\pdfoutput=1

\documentclass[11pt]{article}

\usepackage{EMNLP2023}

\usepackage{times}
\usepackage{latexsym}
\usepackage{paralist}
\usepackage{amsmath}
\usepackage{dsfont}
\usepackage{tikz}
\usetikzlibrary{tikzmark, quotes}
\usepackage{booktabs}
\usepackage{amssymb}
\usepackage{multirow}
\usepackage{arydshln}
\usepackage{color,soul}
\usepackage{amsfonts}
\usepackage{fetamont}
\usepackage[T1]{fontenc}

\newcommand{\rankbox}[1]{%
  \raisebox{0.3ex}{\colorbox{blue!20}{\makebox[5pt]{\textcolor{black}{\small\textbf{#1}}}}}%
}

\newcommand{\overbar}[1]{\mkern 1.5mu\overline{\mkern-1.5mu#1\mkern-1.5mu}\mkern 1.5mu}

\usepackage[T1]{fontenc}

\usepackage[utf8]{inputenc}

\usepackage{microtype}

\usepackage{inconsolata}

\title{Feeding Two Birds or Favoring One? Adequacy–Fluency Tradeoffs in Evaluation and Meta-Evaluation of Machine Translation}

\author{Behzad Shayegh$^{1,*}$ \;\; Jan-Thorsten Peter$^2$ \;\; David Vilar$^2$ \;\; Tobias Domhan$^2$ \\ {\bf Juraj Juraska$^2$ \;\; Markus Freitag$^2$ \;\; Lili Mou$^{1,3}$}
\\
\normalsize $^1$Dept. Computing Science, Alberta Machine Intelligence Institute (Amii), University of Alberta\\
\normalsize $^2$Google\qquad 
$^3$Canada CIFAR AI Chair, Amii\\
\normalsize \texttt{\{the.shayegh, doublepower.mou\}@gmail.com} \\
\normalsize \texttt{\{vilar, jtp, domhant, jjuraska, freitag\}@google.com}
}

\begin{document}
\maketitle
\begingroup
\renewcommand\thefootnote{$\star$}
\footnotetext{Work partially done when the first author was interning at Google.}
\endgroup
\begin{abstract}
We investigate the tradeoff between adequacy and fluency in machine translation. We show the severity of this tradeoff at the evaluation level and analyze where popular metrics fall within it. Essentially, current metrics generally lean toward adequacy, meaning that their scores correlate more strongly with the adequacy of translations than with fluency. More importantly, we find that this tradeoff also persists at the meta-evaluation level, and that the standard WMT meta-evaluation favors adequacy-oriented metrics over fluency-oriented ones. We show that this bias is partially attributed to the composition of the systems included in the meta-evaluation datasets. To control this bias, we propose a method that synthesizes translation systems in meta-evaluation. Our findings highlight the importance of understanding this tradeoff in meta-evaluation and its impact on metric rankings.
\end{abstract}

\section{Introduction}
\label{intro}

As translation systems become more sophisticated and widely adopted, the critical challenge of accurately evaluating their performance has come to the forefront, driving significant ongoing research to the developement of more accurate and robust metrics~\citep{Chatzikoumi_2020, 10.1162/tacl_a_00683}. Traditionally, translation evaluation relied on lexical metrics such as BLEU~\citep{papineni-etal-2002-bleu} and ChrF~\citep{popovic-2015-chrf}, which primarily consider the n-gram overlap between the candidate and reference translations. However, these metrics have been shown insufficient for measuring the quality of modern translation systems~\citep{babych2008sensitivity, freitag-etal-2022-results}. In the recent decade, researchers have been exploring the idea of training neural models to measure translation quality~\citep{wieting2019bleutrainingneuralmachine, ma2019results, freitag-etal-2022-results, 10.1162/tacl_a_00683}, with popular examples including MetricX~\citep{juraska-etal-2023-metricx, juraska-etal-2024-metricx} and Comet~\citep{rei-etal-2020-comet, rei-etal-2022-comet}.

Two key aspects of translation quality, long discussed in the community~\citep{nap1966language, white-oconnell-1993-evaluation, 7045551, martindale-carpuat-2018-fluency}, are\footnote{Different namings and slightly different definitions are used in the literature for these two aspects; however, the main idea remains consistent.}:
\begin{compactitem}
    \item \textit{Fluency}: the grammatical correctness and naturalness of the translation; and
    \item \textit{Adequacy}: how well the translation conveys the meaning of the source text.
\end{compactitem}
\citet{flamich2025feedbirdsscoreaccuracynaturalness} demonstrate that an adequacy--fluency tradeoff exists in translation: optimizing for one aspect eventually sacrifices the other. They further introduce measurements to study the severity of this tradeoff at \textbf{the level of translation systems}.

The aforementioned tradeoff naturally results in an adequacy--fluency tradeoff at \textbf{the level of evaluation metrics}, if we limit our evaluation to a single score: increasing the capability of measuring one aspect eventually comes at the cost of decreasing the capability of measuring the other. It is important to understand this tradeoff and know the position of each metric to avoid undesired biases when optimizing translation systems according to the metric.

\begin{table}
    \centering
    \begin{tabular}{l|c|c}
        \toprule
        Evaluation Set & Concordance & Discordance \\
        \midrule
        En--De'23 & 49 (74\%) & 17 (26\%) \\
        Zh--En'23 & 70 (67\%) & 35 (33\%) \\
        En--De'24 & 98 (72\%) & 38 (28\%) \\
        En--Es'24 & 55 (71\%) & 23 (29\%) \\
        Ja--Zh'24 & 58 (74\%) & 20 (26\%) \\
        \bottomrule
    \end{tabular}
    \caption{Concordance and discordance between adequacy and fluency in MQM datasets~(\S\ref{sec:expdataset}), reported as system pair counts and percentages. Concordance means one system in a pair outperforms the other in both metrics; discordance indicates inconsistent performance across metrics. Notice that 26--33\% discordance, although being the minority, is way far from being negligible and imposes a significant tradeoff when the metric quality improves.}
    \label{tab:Concordance}
    \vspace{-10pt}
\end{table}

In this work, we demonstrate the severity of the adequacy--fluency tradeoff at the evaluation level. We analyze current evaluation setups in WMT~\citep{callison-burch-etal-2007-meta, 10.1162/coli_a_00537}, and show that there is a significant disagreement between adequacy and fluency when ranking systems (Table~\ref{tab:Concordance}). This directly imposes a severe tradeoff, as a metric can only agree with either adequacy or fluency when comparing discordant system pairs. Subsequently, we empirically analyze several contemporary translation metrics to illustrate their positions within this tradeoff.

More importantly, we show that this adequacy--fluency tradeoff extends even to \textbf{the level of meta-evaluation} (the evaluation of evaluation metrics). Meta-evaluation typically compares the scores of a set of candidates given by a metric to those given by humans~\citep{callison-burch-etal-2007-meta}. We show that the selection of the candidates used during the meta-evaluation significantly influences the identified optimal metric within the tradeoff: if there is a higher variance in the candidates' adequacy than in their fluency, the meta-evaluation will lean toward favoring metrics that prioritize adequacy. Conversely, if the fluency of these candidates shows higher variance, the meta-evaluation will prefer fluency-oriented metrics. This can inadvertently guide the development of evaluation metrics, and thus the development of translation systems, toward a particular bias toward either adequacy or fluency. We propose to synthesize candidates with desired variance in their adequacy and fluency scores to conduct a controlled~(balanced) meta-evaluation.

\section{Background \& Related Work}

\subsection{Translation Metrics}

The evaluation of translation systems is a critical aspect of their development and deployment, and has been researched for years~\citep{tao2018ruber, Chatzikoumi_2020}. This research has resulted in the development of a variety of metrics~\citep{mtevalmetricssurvey, 10.1162/tacl_a_00683}.

Traditional metrics, such as BLEU \citep{papineni-etal-2002-bleu}, METEOR \citep{lavie2009meteor}, and ChrF \citep{popovic-2015-chrf}, compare the output of a translation system against one or more human-created reference translations, only at the surface level (e.g., based on string overlap). While widely adopted for their simplicity, these surface-level metrics are shown to be incapable of measuring the quality of modern, high-quality translation systems~\cite{freitag-etal-2022-results}.

Recent advancements have led to the development of \textit{trained metrics}, which leverage machine learning models to assess translation quality~\citep{wieting2019bleutrainingneuralmachine, freitag-etal-2022-results}. These metrics, such as MetricX~\citep{juraska-etal-2023-metricx,juraska-etal-2024-metricx} and Comet~\citep{rei-etal-2020-comet, rei-etal-2022-comet}, are often trained on large datasets of human annotations, allowing them to learn more nuanced correlations between system outputs and perceived quality.

Translation metrics can be divided into two categories: \textit{reference-based} and \textit{reference-free}. Reference-based metrics, such as those mentioned previously, rely on access to a trusted reference translation and compare the candidate against the reference to assess quality. Reference-free metrics, also known as Quality Estimation (QE) metrics, directly evaluate the candidate given the source text, without requiring a reference translation~\citep{ito2025reference}. This capability is particularly valuable in scenarios where human references are scarce or impractical to obtain, offering a more flexible and potentially more accurate evaluation paradigm for contemporary translation systems~\citep{agrawal-etal-2021-assessing}. MetricX and Comet offer their reference-free versions: MetricX QE~\citep{juraska-etal-2023-metricx, juraska-etal-2024-metricx} and CometKiwi~\citep{rei-etal-2022-cometkiwi}.

In our work, we empirically analyze several contemporary translation metrics to illustrate their positions within the adequacy--fluency tradeoff, providing insights into their strengths and limitations.

\subsection{Meta-Evaluation}
\label{sec:bgmetametrics}

Meta-evaluation assesses how well a translation evaluation metric correlates with human annotations~\citep{callison-burch-etal-2007-meta, machacek-bojar-2013-results}. This is traditionally accomplished by computing Pearson, Kendall, and Spearman correlation coefficients~\citep{machacek-bojar-2013-results, mathur-etal-2020-results, freitag-etal-2023-results}, at either the segment\footnote{Following the common terminology in the community, we refer to each data sample as a \textit{segment}.} level~(for individual translation scores) or the system level~(for scores averaged per system).
More recently, pairwise accuracy~\citep[PA;][]{kocmi-etal-2021-ship} and soft pairwise accuracy~\citep[SPA;][]{thompson-etal-2024-improving} have gained prominence as they directly assess how well a metric aligns with human preferences over pairs of translations~\citep{mathur-etal-2020-results}. Given their popularity, we utilize these two metrics for our analysis and elaborate their formulations below.

\paragraph{Pairwise accuracy.} PA assesses whether a metric correctly identifies the better translation (segment level) or the better system (system level) given a pair. Let $m$ and $h$ be the metric and human scores, respectively. PA is formulated as
\begin{align}
\frac{1}{|\mathcal{P}|} \sum_{i,j \in \mathcal{P}} \mathds{1}[\operatorname{sgn}(m_j-m_i) = \operatorname{sgn}(h_j-h_i)]
\label{eqn:PA}
\end{align}
where $\mathcal{P}$ is the set of pairs, and $\operatorname{sgn}$ is the sign function. Pairs with tied human scores are excluded~\citep{kocmi-etal-2021-ship}.

A key challenge for PA at the segment level is to handle tied metric scores. While techniques like tie calibration are employed to address this issue~\citep{deutsch-etal-2023-ties}, recent studies~\citep{perrella-etal-2024-guardians} indicate that such approaches are not reliable and the matter warrants further investigation. This challenge is minor at the system level, since averaging scores over many translations rarely yields ties. We therefore focus on system-level PA in our work.

\label{sec:SPA}
\paragraph{Soft pairwise accuracy.} SPA is a system-level meta-metric that extends PA by incorporating statistical significance from both human and metric scores. Unlike PA's binary agreement, SPA compares $p$-values from hypothesis tests to determine if one system is statistically superior. The SPA score is formulated as
\begin{align}
    \frac{1}{|\mathcal{P}|} \sum_{i,j \in \mathcal{P}} (1-|p(h_j>h_i) - p(m_j>m_i)|)
\end{align}
Here, $p(x_j>x_i)$ is the $p$-value from a permutation test~\citep{65dc97ac-df33-39b9-8741-82976479818f} assessing if system $j$ is superior to system $i$ based on scores $x$.
SPA is sensitive to the magnitude of the score differences, reflecting the metric's confidence in its preference, and compares this confidence against that of human judgments.

\section{Experimental Setup}

\subsection{Dataset}
\label{sec:expdataset}

The Multidimensional Quality Metrics (MQM) framework provides a standardized and granular guideline for characterizing and classifying errors in translations~\citep{mariana2015multidimensional}. Instead of being a score-based annotation, the MQM framework annotates translation errors. Each error is annotated as a substring of the translation with a category (such as accuracy, fluency, terminology, and style) and a severity level (usually major, minor, and neutral). This framework enables a diagnostic understanding of translation system performance, as well as more reliable and standardized evaluation scores.

WMT adopts MQM annotations~\citep{freitag-etal-2021-experts} for human evaluation, producing datasets of detailed translation error annotations. We use this dataset of years 2023 and 2024 in our meta-evaluation and analysis. We reserve data from prior years (2020--2022) for training to avoid leakage. Our setup is consistent with practice in previous studies~\citep{juraska-etal-2024-metricx, rei-etal-2022-comet} for fair comparison.

It is common to transform MQM annotations into a single score~\citep{freitag-etal-2022-results, freitag-etal-2023-results}. This involves assigning a specific penalty to each error based on its category and severity~\citep{freitag-etal-2021-results}. These penalties are then summed across all errors in the translation. We refer to this aggregated value as the \textit{All~MQM} score, where a lower score indicates higher quality.

For a more granular analysis, \citet{flamich2025feedbirdsscoreaccuracynaturalness} categorize MQM errors into adequacy and fluency classes~(Tables~\ref{tab:mqm_error_classification} and~\ref{tab:mqm_error_classification_enes} in Appendix~\ref{appendix:error-classification}), resulting in \textit{Adequacy~MQM} and \textit{Fluency~MQM} scores for each segment. We use these specialized scores for our tradeoff analysis. We present detailed statistics for our dataset in Appendix~\ref{appendix:dataset_statistics}.

\subsection{Meta-Metrics}
\label{sec:metametrics}

To study the performance of evaluation metrics in the adequacy--fluency tradeoff, we separately assess their ability to measure translation adequacy and fluency. To achieve this, we compute meta-metrics---namely, PA and SPA~(\S\ref{sec:bgmetametrics})---with respect to Adequacy~MQM and Fluency~MQM. In Appendix~\ref{sec:pavsspa}, we point out the necessity of including both PA and SPA in such studies due to their potentially different behavior in measuring bias toward adequacy or fluency.

\subsection{Translation Metrics in Our Analysis}
\label{sec:underanalmetrics}

In our work, we select a set of diverse and widely used metrics to analyze their positions within the adequacy--fluency tradeoff. These metrics include \textbf{BLEU}~\citep{papineni-etal-2002-bleu} and \textbf{ChrF}~\citep{popovic-2015-chrf}, representing overlap-based metrics; \textbf{MetricX}~\citep{juraska-etal-2024-metricx} and \textbf{Comet}~\citep{rei-etal-2022-comet}, representing trained metrics; and \textbf{MetricX~QE} and \textbf{CometKiwi}~\cite{rei-etal-2023-scaling}, representing reference-free metrics. 

Moreover, we introduce a set of extreme translation metrics designed to measure only adequacy or fluency, which will provide a clearer view of the tradeoff between the two. Specifically, we develop \textbf{AdequacyX} and its reference-free variant, \textbf{AdequacyX~QE}, trained exclusively on MQM errors categorized under adequacy. Likewise, we develop \textbf{FluencyX} trained on fluency annotations.

Technically, these trainable metrics adopt the same neural architecture as MetricX~\cite{juraska-etal-2024-metricx}, and are fine-tuned from an mT5 checkpoint~\citep{xue-etal-2021-mt5}. Following \citet{juraska-etal-2024-metricx}, we train these models on MQM~annotations~(WMT 2022), covering En--De, En--Ru, and En--Zh.\footnote{We limit En--Zh to the conversation, e-commerce, and social domains, in line with~\citet{juraska-etal-2024-metricx}.} AdequacyX and AdequacyX~QE are trained to predict Adequacy~MQM, while FluencyX is trained to predict Fluency~MQM. Unlike \citet{juraska-etal-2024-metricx}, we exclusively use MQM annotations for fine-tuning, omitting direct assessments (DA) and synthetic data, as these do not offer a clear adequacy–fluency distinction.

It should be noted that AdequacyX~QE does not take the reference translation as input as it is a reference-free metric. We further restrict FluencyX to the candidate translation alone (no source or reference), similar to SENTINEL\textsubscript{CAND} in \citet{perrella-etal-2024-guardians}, given that fluency is considered source-independent.

\begin{table*}[t]
    \centering
    \begin{tabular}{l|r r r r|r r r r | r}
    \toprule
         & \multicolumn{4}{c|}{Adequacy MQM} & \multicolumn{4}{c|}{Fluency MQM} & \\
          & Variance & F-stat & PA & SPA & Variance & F-stat & PA & SPA & $B(\Delta p)$ \\
        \midrule
        En--De'23 & $0.31$ & $36.5$ & $0.98$ & $0.97$ & $0.06$ & $7.0$ & $0.76$ & $0.75$ & $0.08^A$ \\
        Zh--En'23 & $0.23$ & $80.6$  & $0.97$ & $0.93$ & $0.03$ & $12.9$ & $0.70$ & $0.74$ & $0.03^A$\\
        En--De'24 & $0.24$ & $13.7$& $0.88$ & $0.87$ & $0.11$ & $9.5$ & $0.85$ & $0.81$ & $0.04^A$ \\
        En--Es'24 & $0.61$ & $26.4$ & $0.96$ & $0.94$ & $0.28$ & $4.6$ & $0.74$ & $0.77$ & $0.13^A$ \\
        Ja--Zh'24 & $0.40$ & $35.3$ & $1.00$ & $0.98$ & $0.10$ & $4.8$ & $0.74$ & $0.75$ & $0.12^A$ \\
        \hline
        Macro-Avg & $0.36$ & $38.5$ & $0.96$ & $0.94$ & $0.12$ & $7.7$ & $0.76$ & $0.76$ \\
        \bottomrule
    \end{tabular}
    \caption{Adequacy~MQM and Fluency~MQM statistics across different evaluation sets. The reported statistics include the variance of system-level scores, the (S)PA of the scores with respect to All~MQM, the F-statistic, and the $B$~transformation of the difference between the $p$-values of Adequacy~MQM and Fluency~MQM. The last two are explained in~\S\ref{sec:measuringimplicit}.}
    \label{tab:stds}
    \vspace{-12pt}
\end{table*}

To have a fair comparison, we also train our own MetricX and MetricX~QE variants; they are identical to AdequacyX~(QE), except that they predict All~MQM scores. The key difference between our variants and the original MetricX~(QE) is the exclusion of DA and synthetic data.

Additionally, we include the log-perplexities of a pretrained Gemma model as a fluency measure, following \citet{flamich2025feedbirdsscoreaccuracynaturalness}. We use \textbf{Gemma~3~4B}~\citep{team2025gemma} in this work.

\section{Analysis: Tradeoff at the Meta-Evaluation Level}
\label{sec:imbalance}

In this section, we explore the adequacy--fluency tradeoff at the meta-evaluation level. We first discuss in~\S\ref{sec:understandingimbalance} {how the tradeoff exists in meta-evaluation, and show that the imbalance in WMT meta-evaluation datasets imposes a bias that favors adequacy-oriented metrics.} We argue that this bias consists of both an intrinsic and an extrinsic component, and that we aim to control the latter. In~\S\ref{sec:measuringimplicit}, we design a metric to quantify the extrinsic bias, and in~\S\ref{sec:reducingimplicit}, we propose a practical approach to reduce it. Finally we demonstrate in~\S\ref{sec:metaevaluationbiasimportance} the impact of meta-evaluation bias on the ranking of translation metrics by comparing this ranking before and after reducing the meta-evaluation extrinsic bias.

\subsection{{Understanding the Meta-Evaluation Bias}}
\label{sec:understandingimbalance}

Translation meta-evaluation compares the ranking of translation systems produced by a given metric against the ranking based on human annotations, which in our case is the All~MQM score. In this section, we will show that the All~MQM ranking reflects systems' adequacy more than their fluency, leading the WMT meta-evaluation to favor adequacy-oriented metrics.

For simplicity, we assume that
$\text{All~MQM} = \text{Adequacy~MQM} + \text{Fluency~MQM}$.\footnote{Although there are some errors that are not categorized as either adequacy or fluency, they are rare and have minor influence; therefore, we omit them for simplicity.} Therefore, the ranking by All~MQM {(thus the meta-evaluation assessment)} is more influenced by the component with higher variance, {and we say the meta-evaluation is \textit{biased} toward that component}.

As shown in Table~\ref{tab:stds}, Adequacy~MQM exhibits a much higher variance than Fluency~MQM {in system level scores, consequently having a greater influence on the All~MQM system ranking} {and the meta-evaluation {system-level assessment}}. This influence is further demonstrated by the stronger alignment between All~MQM and Adequacy~MQM in terms of both PA and SPA. The question now is:
\textit{Should we embrace this dominance of adequacy, or is it a bias we should mitigate?}

Answering the above question requires noticing that the higher variance of the system-level Adequacy~MQM scores, and consequently its greater influence, can be attributed to two possible causes:
\begin{compactitem}
\item \textit{Intrinsic variation,} {which is due to the preference of the MQM framework and annotators}. For example, adequacy errors may be generally considered more severe, leading to larger assigned penalties and thus greater variance.
\item \textit{Extrinsic variation,} {which is due to the choice of translation systems during meta-evaluation. For example, we may} select systems that differ primarily in adequacy, while performing similarly in fluency. This naturally leads to higher system-level variance in Adequacy~MQM {than in} Fluency~MQM.
\end{compactitem}
Both of the above factors contribute to the variances of system-level Adequacy~MQM and Fluency~MQM, which in turn lead to the bias of the meta-evaluation. Therefore, we say that the meta-evaluation bias consists of two components: \textit{intrinsic bias} and \textit{extrinsic bias}.

{The intrinsic bias reflects translation experts’ beliefs about translation errors. Therefore, we retain intrinsic bias in this study. However, whether the extrinsic bias should be eliminated, retained, or partially kept is debatable and depends on the application. In our study, we aim to avoid it in order to gain a clearer understanding of the adequacy--fluency tradeoff at the evaluation level. In other contexts, one could argue that adequacy assessment should have greater influence on meta-evaluation outcomes, even at the cost of reduced fluency influence. In both cases, the extrinsic bias must first be measured separately from the intrinsic bias.}

\subsection{{Measuring the Extrinsic Bias}}
\label{sec:measuringimplicit}

To study the extrinsic bias, we propose to compare the F-statistics~\citep{d9405b68-bf4d-3925-a457-0f7aeae8e78d} for Adequacy~MQM and Fluency~MQM. In each case, the F-statistic is formulated as follows:
\begin{align}
\text{F-statistic} = \frac{\text{between-system variation}}{\text{within-system variation}}
\label{eqn:fstat}
\end{align}
where the numerator and denominator are
\begin{align}
    \parbox{2.5cm}{between-system\\variation} &= \sum_{i=1}^K \frac{N \cdot (\bar{s}_i - \bar{s})^2}{K-1} \\
    \parbox{2.5cm}{within-system\\variation} &= \sum_{i=1}^K \sum_{j=1}^{N} \frac{N \cdot (s_{i,j} - \bar{s}_i)^2}{N-K}
\end{align}
In our scenario, $s_{i,j}$ is the Adequacy~MQM or Fluency~MQM score of the candidate translation given by the $i$th system for the $j$th segment, $\bar{s}_i$ is the average of all the scores by the $i$th system, and $\bar{s}$ is the average of all the scores. $N$ and $K$ are the numbers of segments and systems, respectively.

The F-statistic is effective at distinguishing extrinsic variation from intrinsic variation because of how its numerator and denominator are defined. On one hand, the intrinsic variation can be regarded as a multiplicative constant within the $s$ variables; it appears in both the numerator and denominator of Eqn.~\eqref{eqn:fstat} and thus cancels out. On the other hand, the F-statistic captures extrinsic variation, which is due to the selected systems in the meta-evaluation. It does so by normalizing between-system variation with within-system variation.

\begin{table*}
    \centering
    \begin{tabular}{r | ccc | rrr | rrr | r}
        \toprule
        & {Original} & \multicolumn{2}{c|}{Synthesized by} & \multicolumn{3}{c|}{{Adequacy~MQM}} & \multicolumn{3}{c|}{{Fluency~MQM}} & \\
        & & Adequacy & Fluency & F-stat & PA & SPA & F-stat & PA & SPA & $B(\overbar{\Delta p})$ \\
        \midrule
        1 & \checkmark & -- & -- & $38.5$ & $0.96$ & $0.94$ & $7.7$ & $0.76$ & $0.76$ & $0.120^A$ \\
        2 & -- & \checkmark & -- & $213.3$ & $0.94$ & $0.95$ & $2.5$ & $0.53$ & $0.54$ & $0.590^A$ \\
        3 & -- & -- & \checkmark & $7.5$ & $0.62$ & $0.71$ & $165.1$ & $0.82$ & $0.78$ & $0.450^F$ \\
        4 & \checkmark & \checkmark & -- & $111.9$ & $0.95$ & $0.95$ & $4.9$ & $0.61$ & $0.62$ & $0.130^A$ \\
        5 & \checkmark & -- & \checkmark & $21.7$ & $0.80$ & $0.83$ & $75.5$ & $0.73$ & $0.73$ & $0.030^F$ \\
        6 & -- & \checkmark & \checkmark & ${93.6}$ & $0.84$ & $0.87$ & ${72.4}$ & $0.59$ & $0.60$ & $\bf{{0.005}^A}$ \\
        7 & \checkmark & \checkmark & \checkmark & $72.7$ & $0.88$ & $0.88$ & $48.8$ & $0.63$ & $0.64$ & $\bf{0.005^A}$ \\
        \bottomrule
    \end{tabular}
    \caption{Adequacy~MQM and Fluency~MQM statistics for different meta-evaluation setups, along with the $B$~transformation of the difference between their corresponding $p$-values, macro-averaged across evaluation sets. Each meta-evaluation setup includes one or more system sets drawn from the original, synthesized-by-adequacy, and synthesized-by-fluency system sets.
}
    \label{tab:synthesized_systems}
    \vspace{-10pt}
\end{table*}

Notice that the F-statistic depends on~$N$ and~$K$, which makes it difficult to interpret. To this end, we adopt the ANOVA framework~\citep{heiman2001understanding} and use $p$-values to quantify extrinsic variation. We perform this analysis separately for Adequacy~MQM and Fluency~MQM.

We assume that, within each translation system, scores are independent and normally distributed around that system’s mean, and that all systems share the same variance\footnote{ANOVA is robust to moderate variance inequality, so this assumption need only hold approximately~\citep{lowry2014concepts}. For completeness, we report in Appendix~\ref{sec:alteranova} the results using the variance-inequality-tolerant alternative ANOVA~\citep{c0eb885e-c393-3777-9bf1-621bcb49b979}, which yield the same conclusions; We prefer the standard approach for its simplicity and greater numerical stability.}; i.e., $s_{i,j} \overset{\mathrm{iid}}{\sim} \mathcal{N}(\mu_i, \sigma^2)$. Under the null hypothesis that all systems have the same mean ($\mu_1 = \mu_2 = \cdots$), the F-statistic in Eqn.~\eqref{eqn:fstat} follows an F distribution~\cite{johnson1995continuous} with degrees of freedom $K-1$ and $N-K$. The right-tail $p$-values are then given by $1 - \operatorname{CDF}(\text{F-statistic})$, where $\operatorname{CDF}$ denotes the cumulative distribution function of the F-distribution. Converting F-statistics to $p$-values puts results on a common probability scale, and makes the interpretation independent of $N$ and $K$. We use this $p$-value as our measure of extrinsic variation.

As mentioned in \S\ref{sec:understandingimbalance}, extrinsic variations of Adequacy~MQM and Fluency~MQM contribute to their respective variances. An asymmetry in these variations induces a bias of meta-evaluation toward adequacy or fluency, which we term \textit{extrinsic bias}. We quantify this bias by
\begin{align}
    \Delta p = p_\text{\;Adequacy MQM} - p_\text{\;Fluency MQM}
\end{align}
where $p_\text{\;Adequacy MQM}$ and $p_\text{\;Fluency MQM}$ are the respective $p$-values. A positive $\Delta p$ indicates bias toward adequacy, a negative $\Delta p$ indicates bias toward fluency, and $\Delta p = 0$ indicates no bias between adequacy and fluency.

We observe that the range of $\Delta p$ is typically very small (from $10^{-7}$ to $10^{-31}$ in Table~\ref{tab:stds}), even in cases where severe bias is present (as shown in our later analysis). For presentation purposes, we report the degree of bias using $B(\Delta p) = \frac{1}{1 - \log |\Delta p|}$, and append a special symbol to indicate which aspect dominates: $A$ for adequacy ($\Delta p>0$) and $F$ for fluency ($\Delta p<0$). Note that $B \in [0,1]$, with a lower $B$ indicating less bias.

We report the $B$ values in Table~\ref{tab:stds}, where we compare the influence of Adequacy~MQM and Fluency~MQM on the meta-evaluation in five evaluation datasets. Results show a consistent extrinsic bias toward adequacy. We argue that this behavior stems from the particular composition of translation systems in the WMT datasets, and may be controlled by carefully selecting or synthesizing systems with more variance in their Fluency~MQM.

\subsection{Reducing the Extrinsic Bias}
\label{sec:reducingimplicit}

In this subsection, we provide a method to reduce meta-evaluation extrinsic bias. This not only has practical values, but also allows us to better evaluate the adequacy--fluency bias of existing translation evaluation metrics (to be discussed in~\S\ref{sec:exptradeoffeval}).

We accomplish this by considering additional pseudo-translation systems in the meta-evaluation. In particular, we synthesize two sets of translation systems: one exhibiting extreme variations in system-level Adequacy~MQM, and the other in Fluency~MQM. This design helps to control the variation in each dimension and thereby reduces extrinsic bias through controlled mixing of candidate translation systems.

Suppose we have $K$ original translation systems. We synthesize $K$ adequacy-oriented systems in the following way: for each segment, we assume that the $k$th synthesized system generates the $k$th most adequate translation (according to Adequacy~MQM), among the original $K$ translation systems' outputs.\footnote{For segments with tied scores, we rank them randomly.} It is easy to see that, given the available translation candidates, such $K$ synthesized translate systems exhibit extreme variation in Adequacy~MQM at the system level. Likewise, we synthesize $K$ fluency-oriented systems, and in total, we have $3K$ translation systems for meta-evaluation. It is emphasized that our research does not require additional annotation effort, as the above synthesizing process utilizes readily available MQM scores.

Table~\ref{tab:synthesized_systems} illustrates how synthesized systems help control the extrinsic bias. The table reports the macro-average of metrics across the five evaluation sets~(\S\ref{sec:expdataset}).

Rows~2 and~3 represent extreme configurations that maximize the F-statistic for Adequacy~MQM and Fluency~MQM, respectively. We see that they yield extreme $B$ values, with Row~2 being biased toward adequacy and Row~3 toward fluency. This expected pattern confirms that (1)~the $B$ value effectively captures extrinsic bias arising from system selection, and (2) our synthesis method can control such a bias.

We observe that the setup in Row~5 has an overall bias toward adequacy (indicated by higher PA and SPA scores), although the extrinsic bias is toward fluency (indicated by the $F$ annotation).
This suggests the existence of intrinsic bias discussed in \S\ref{sec:understandingimbalance}, further justifying the need of separately quantifying the intrinsic and extrinsic bias.

We also observe that Rows~6 and~7 yield the lowest~${B}$~values, and we consider them the most balanced meta-evaluation setups, which will be used in our study of adequacy--fluency tradeoff at the evaluation level in \S\ref{sec:exptradeoffeval}.

\subsection{The Effect of Meta-Evaluation Bias on Metric Comparisons}
\label{sec:metaevaluationbiasimportance}

In the previous parts, we have demonstrated that the standard WMT meta-evaluation is biased toward adequacy, and we propose to carefully synthesize pseudo-translation systems to control this bias. In this part, we show how the meta-evaluation bias can influence the development of translation metrics.

Table~\ref{tab:orders} presents how the original WMT setup (Row~1 of Table~\ref{tab:synthesized_systems}) and \textit{our balanced setup} (Row~6 of Table~\ref{tab:synthesized_systems}) rank various metrics, using PA and SPA as the meta-metrics.

It is interesting to examine ``CometKiwi~22~XXL'' and ``MetricX~(ours)'' in detail, shown by the bold lines in Table~\ref{tab:orders}. As seen, CometKiwi~22~XXL consistently and significantly outperforms MetricX~(ours) under the original meta-evaluation, which is biased toward adequacy; however, this trend is reversed under our balanced meta-evaluation setup. In~\S\ref{sec:exptradeoffeval}, we will show that CometKiwi~22~XXL has a considerable bias toward adequacy, whereas MetricX~(ours) demonstrates a more balanced behavior. This comparison shows that a metric is favored if its bias in the adequacy--fluency spectrum aligns with that of the meta-evaluation; consequently, the development of translation metrics inherits the bias of the meta-evaluation. Our analysis highlights the importance of studying translation meta-evaluation imbalance.

\begin{table}[t]
\centering
\resizebox{\linewidth}{!}{
\begin{tabular}{l | c c | c c}
\toprule
{Metric}
& \multicolumn{2}{c|}{{Original setup}} & \multicolumn{2}{c}{{Our balanced setup}} \\
& PA & SPA & PA & SPA \\
    \midrule
    AdequacyX & 0.881 \rankbox{4} & 0.866 \rankbox{4} & 0.847 \rankbox{1} & 0.833 \rankbox{1} \\
    AdequacyX QE & 0.885 \rankbox{3} & 0.873 \rankbox{1} & 0.834 \rankbox{2} & 0.824 \rankbox{5} \\
    FluencyX & 0.697 \rankbox{13} & 0.720 \rankbox{12} & 0.676 \rankbox{12} & 0.675 \rankbox{13} \\
    \hdashline
    \textbf{MetricX~(ours)} & \textbf{0.862 }\rankbox{7} & \textbf{0.845} \rankbox{8} & \textbf{0.833} \rankbox{3} & \textbf{0.829} \rankbox{3} \\
    MetricX QE (ours) & 0.878 \rankbox{6} & 0.861 \rankbox{6} & 0.809 \rankbox{6} & 0.823 \rankbox{6} \\
    MetricX-24 & 0.880 \rankbox{5} & 0.866 \rankbox{3} & 0.820 \rankbox{5} & 0.831 \rankbox{2} \\
    MetricX-24 QE & 0.888 \rankbox{2} & 0.867 \rankbox{2} & 0.820 \rankbox{4} & 0.827 \rankbox{4} \\
    \hdashline
    Comet 22 & 0.850 \rankbox{8} & 0.847 \rankbox{7} & 0.800 \rankbox{8} & 0.804 \rankbox{8} \\
    CometKiwi 22 & 0.813 \rankbox{9} & 0.802 \rankbox{9} & 0.782 \rankbox{9} & 0.772 \rankbox{9} \\
    \textbf{CometKiwi 22 XXL} & \textbf{0.889} \rankbox{1} & \textbf{0.862} \rankbox{5} & \textbf{0.805} \rankbox{7} & \textbf{0.815} \rankbox{7} \\
    \hdashline
    Gemma 3 (4B) & 0.701 \rankbox{12} & 0.754 \rankbox{10} & 0.652 \rankbox{13} & 0.736 \rankbox{10} \\
    \hdashline
    BLEU (sent. level) & 0.726 \rankbox{11} & 0.712 \rankbox{13} & 0.726 \rankbox{10} & 0.689 \rankbox{12} \\
    ChrF (sent. level) & 0.739 \rankbox{10} & 0.731 \rankbox{11} & 0.722 \rankbox{11} & 0.721 \rankbox{11} \\
    \bottomrule
\end{tabular}
}
\caption{Meta-evaluation of translation metrics using the original WMT setup and our balanced setup. Decimal numbers denote the PA and SPA scores, macro-averaged across our five evaluation sets; blue squares indicate the corresponding rankings. We bold MetricX~(ours) and CometKiwi~22~XXL, as they are discussed in detail in the main text. Note that the results here are not identical to those in the WMT reports as we are reporting the average results across specific evaluation sets.}
\label{tab:orders}
\end{table}

\begin{table}
    \centering
    \begin{tabular}{l|c|c}
        \toprule
        Evaluation Set & Concordance & Discordance \\
        \midrule
        En--De'23 & 136 (49\%) & 140 (51\%) \\
        Zh--En'23 & 124 (29\%) & 311 (71\%) \\
        En--De'24 & 282 (50\%) & 279 (50\%) \\
        En--Es'24 & 139 (43\%) & 186 (57\%) \\
        Ja--Zh'24 & 156 (51\%) & 149 (49\%) \\
        \bottomrule
    \end{tabular}
    \caption{Concordance and discordance between adequacy and fluency in our balanced setup, reported as system pair counts and percentages.}
    \label{tab:ConcordanceBalanced}
    \vspace{-10pt}
\end{table}

\begin{table*}[t]
\begin{center}
\includegraphics[width=\linewidth]{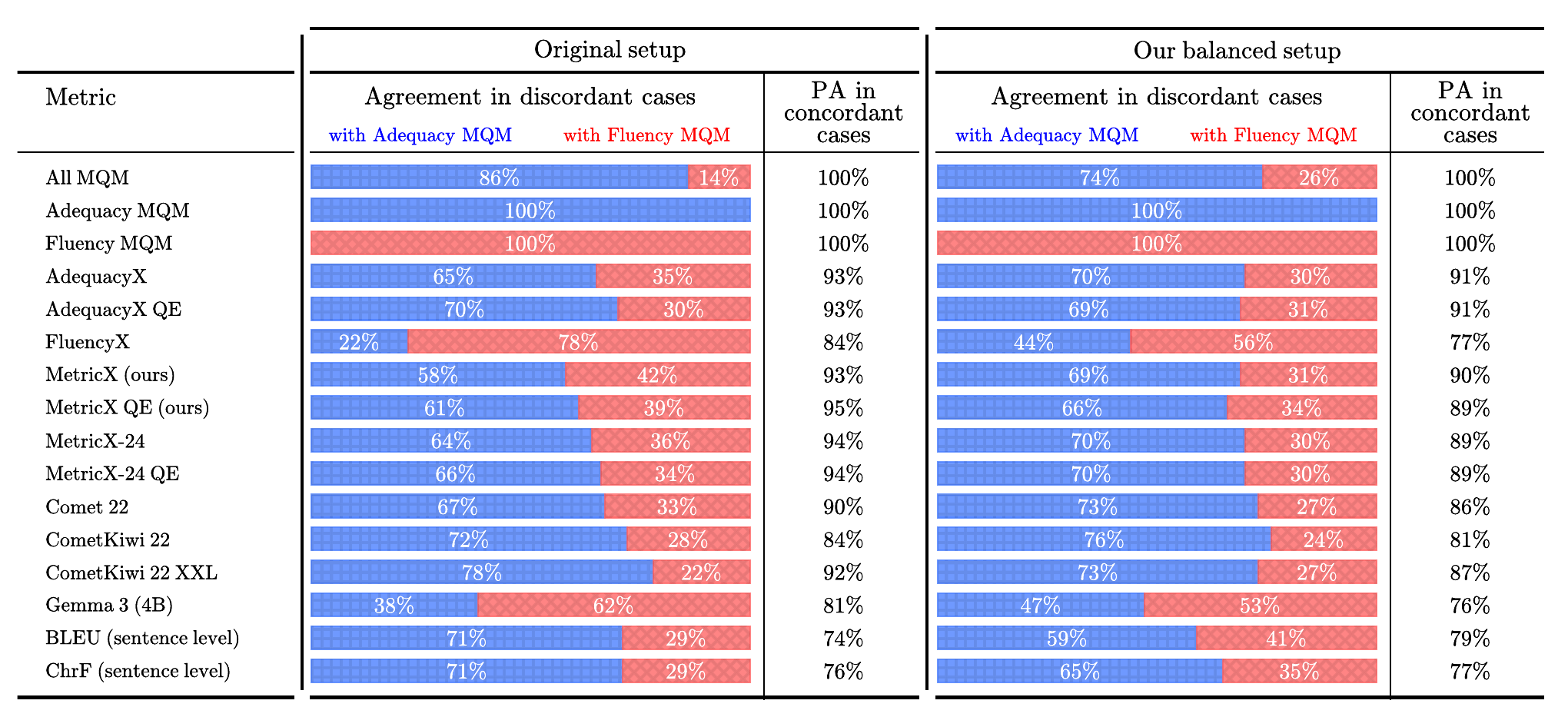}
\end{center}
\vspace{-10pt}
\caption{PA breakdown for the original setup and our balanced setup, macro-averaged across five evaluation sets. For each metric, the agreements with Adequacy~MQM and Fluency~MQM sum to 1, by formulation and because ties between system-level Adequacy~MQM and Fluency~MQM scores are rare.}
\label{fig:PAB}
\vspace{-10pt}
\end{table*}

\section{Analysis: Tradeoff at the Evaluation Level}
\label{sec:exptradeoffeval}

In this section, we analyze the adequacy--fluency tradeoff at the evaluation level and determine the bias of each metric within this tradeoff. Note that the notion of bias considered here is related to but distinct from that in \S\ref{sec:imbalance}. We say a metric is biased toward adequacy or fluency if it ranks translation systems in a way that is more aligned with the ranking derived solely from that aspect, as measured by the corresponding MQM scores.

To study the aforementioned bias, we would like to separately meta-evaluate the adequacy and fluency of each metric, using the original WMT meta-evaluation setup and our balanced setup described in Row~6 of Table~\ref{tab:synthesized_systems}.\footnote{Although Rows~6 and~7 in Table~\ref{tab:synthesized_systems} have similar $B$ values, we prefer the former because it is fully synthesized and has a clearer construction. For completeness, we also report results for the setup in Row~7 in Appendix~\ref{appendix:setup7}.} Notice that the latter is a synthetic setup, which may not be representative of real-world situations. It is included only for our adequacy--fluency analysis.

From the data statistics in Table~\ref{tab:ConcordanceBalanced}, we observe an even stronger disagreement between adequacy- and fluency-based system rankings under the balanced setup than under the original setup (Table~\ref{tab:Concordance}). This is expected, as the balanced setup synthesizes the most- and least-adequate systems, which are unlikely to correspond to the most- and least-fluent systems, and vice~versa.

Despite the significant discordance between Adequacy~MQM and Fluency~MQM, we also observe a large number of concordant cases~(i.e., one system outperforms another in both Adequacy~MQM and Fluency~MQM). This raises a challenge when we separately analyze adequacy and fluency. To address this, we design three experimental protocols, as follows.

\begin{figure*}[t]
{\fontfamily{sfdefault}\selectfont\footnotesize
\qquad\qquad\qquad\qquad\qquad\; Original setup \qquad\qquad\qquad\qquad\qquad\qquad\qquad\quad\; Our balanced setup
\vspace{-4pt}}
\begin{center}
\includegraphics[width=\linewidth]{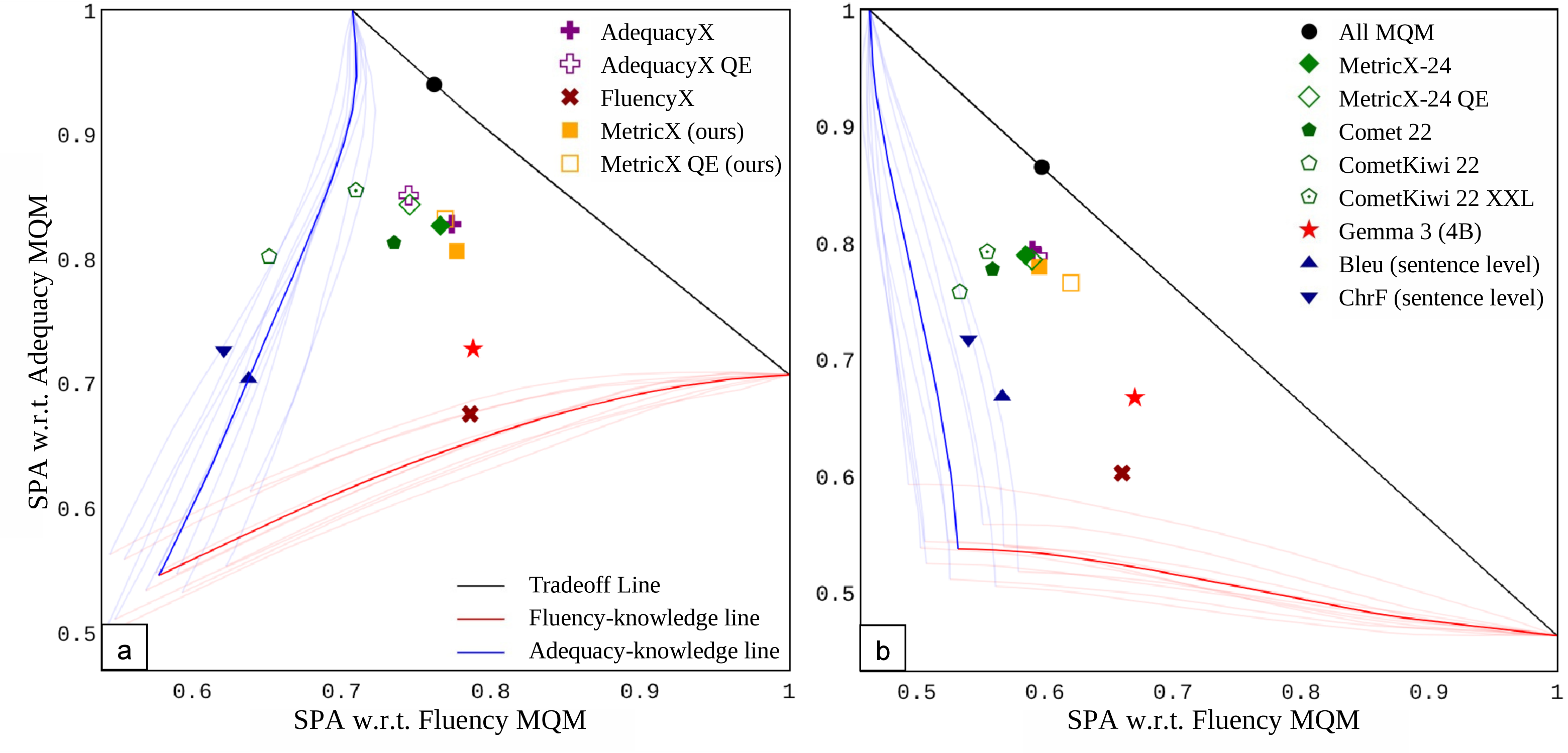}
\end{center}
\vspace{-5pt}
\caption{SPA plane using (a) the original setup and (b) our balanced setup. Legends are shared between the two figures. Each shadow line represents the combination of Adequacy~MQM or Fluency~MQM with a specific random noise instance. The solid red and blue lines are the average of the shadow lines.}
\label{fig:SPAplane}
\vspace{-10pt}
\end{figure*}

\subsection{PA Breakdown}
\label{sec:pab}
In this part, we design an experimental protocol using PA to separately analyze how a metric assesses adequacy and fluency, without being misled by the concordant cases. Since PA counts binary agreements, we may simply exclude concordant cases from our adequacy/fluency analysis. 

Specifically, we first split our system pairs into two disjoint subsets based on whether their Adequacy~MQM and Fluency~MQM are concordant or discordant. Then, for each metric we report:
\begin{compactitem}
    \item PA in concordant cases, measuring the metric's general performance;
    \item PA with respect to Adequacy~MQM in discordant cases, which measures the metric's bias toward adequacy (thus we call it ``agreement with Adequacy~MQM''); and
    \item PA with respect to Fluency~MQM in discordant cases, which is called ``agreement with Fluency MQM.''
\end{compactitem}

In Table~\ref{fig:PAB}, we present the results macro-averaged\footnote{In our preliminary experiments, we also analyzed the micro-averaged results, which led to the same conclusions as those obtained here.} across different evaluation sets.\footnote{We report results for individual evaluation sets under the original setup in Appendix~\ref{appendix:pereval}, where we observe diverse behaviors by the metrics. This variation may be due to the noise of metric performance, which tends to be smoothed out when averaged across evaluation sets.}

As seen, both BLEU and ChrF achieve more agreement with Adequacy~MQM than Fluency~MQM. This provides empirical evidence for the belief in the literature that BLEU and ChrF lean toward adequacy~\cite{flamich2025feedbirdsscoreaccuracynaturalness}. On the other hand, FluencyX and Gemma~3 exhibit a stronger bias toward fluency, which aligns with their design. Moreover, All~MQM shows a strong alignment with Adequacy~MQM in the original setup and a weaker alignment in our balanced setup, confirming our claims in~\S\ref{sec:imbalance} that current WMT meta-evaluation is biased toward adequacy and that such bias is partially mitigated in our balance meta-evaluation setup.

Interestingly, FluencyX and Gemma~3 show some alignment with adequacy, despite lacking access to the source texts. This can be explained by two possibilities: (1) These metrics are imperfect and fail to align entirely with Fluency~MQM, although designed to do so. (2) When the metrics are trained, they have learned strong prior for adequacy errors (e.g., due to the adequacy--fluency correlation of training segments). This aligns with the findings of \citet{perrella-etal-2024-guardians}, who show that metrics based solely on candidate translations or source texts can achieve high performance.

For most other metrics, we find that they exhibit stronger alignment with Adequacy~MQM than with Fluency~MQM. Among them, the MetricX variants display a relatively more balanced behavior than other metrics, including the Comet variants.

Finally, we would like to point out that our analysis  sheds light on the position of different evaluation metrics within the adequacy--fluency tradeoff. However, whether a metric should achieve 50\%--50\% balance or mimicking All MQM (which is highly biased toward adequacy but often treated as the evaluation ground truth) is a task-specific design choice.

\subsection{SPA Plane}
\label{sec:spa_pane}
We also aim to conduct an adequacy–fluency analysis using SPA. However, we cannot mirror the PA breakdown in \S\ref{sec:pab}, because SPA lacks a binary notion of concordance versus discordance. Instead, we perform qualitative visualization to demonstrate the adequacy--fluency tradeoff of a metric.

Specifically, we design a plot where the x-axis and y-axis represent SPA computed with respect to Fluency MQM and Adequacy~MQM, respectively. Each metric is shown as a single point in this space. We then augment the plot with three sentinel lines:
\begin{compactitem}
    \item \textit{Tradeoff line} (the black lines in Figure~\ref{fig:SPAplane}), representing the linear interpolation of Adequacy~MQM and Fluency~MQM. No system can surpass the tradeoff line.
    \item \textit{Adequacy-knowledge line} (the blue lines in Figure~\ref{fig:SPAplane}), derived from linear interpolation of Adequacy~MQM and random scores (i.e., uniform in the range of Adequacy MQM scores for each segment). The interpolation is accomplished by weighted average of the scores. 
    Points on this line achieve fluency scores solely due to the correlation between Adequacy~MQM and Fluency~MQM.
    \item \textit{Fluency-knowledge line} (the red lines in Figure~\ref{fig:SPAplane}), derived from linear interpolation of Fluency~MQM and the random score.
\end{compactitem}
The latter two lines are averaged from 10 computations with different random score instances.

These three lines form a triangle-like shape with vertices at Adequacy~MQM, Fluency~MQM, and pure random noise. A metric's closeness to the tradeoff line indicates its general quality, while its closeness to the other two lines reveals its bias toward adequacy or fluency.

Figure~\ref{fig:SPAplane} presents the results, also macro-averaged across the evaluation sets in \S\ref{sec:pab}. The black line illustrates the severity of the adequacy--fluency tradeoff in translation evaluation (given a certain meta-evaluation setup): a larger top-right area indicates a more severe tradeoff, as this area is not reachable. As shown, our balanced setup exhibits a more severe tradeoff, which is consistent with Table~\ref{tab:ConcordanceBalanced}. Moreover, our balanced setup has a larger angle between the blue and red lines, indicating a lower correlation between the two aspects.

Regarding specific metrics, we observe that FluencyX and Gemma~3 lie close to the fluency boundary (red line), while other metrics (namely, BLEU, ChrF, Comet, and MetricX) are biased toward adequacy. In particular, Comet variants are more biased than MetricX variants. The SPA results are consistent with the analysis using PA~(\S\ref{sec:pab}).

\begin{table}[t]
    \centering
\resizebox{\linewidth}{!}{
    \begin{tabular}{l || c | c || c | c}
    \toprule
         & \multicolumn{2}{c||}{Unnormalized sensitivity} & \multicolumn{2}{c}{Normalized sensitivity} \\
        Metric & Adequacy & Fluency & Adequacy & Fluency \\
        \midrule
        All~MQM & 0.995 & 0.998 & 0.882 & 0.353 \\
        Adequacy~MQM & 1.000 & 0.000 & 1.000 & 0.000 \\
        Fluency~MQM & 0.000 & 1.000 & 0.000 & 1.000 \\
    \hdashline
        AdequacyX & 0.211 & 0.146 & 0.667 & 0.184 \\
        AdequacyX QE & 0.143 & 0.084 & 0.474 & 0.111 \\
        FluencyX & 0.010 & 0.018 & 0.202 & 0.145 \\
    \hdashline
        MetricX~(ours) & 0.125 & 0.100 & 0.644 & 0.206 \\
        MetricX QE~(ours) & 0.081 & 0.070 & 0.521 & 0.180 \\
        MetricX-24 & 0.379 & 0.257 & 0.591 & 0.160 \\
        MetricX-24 QE & 0.300 & 0.207 & 0.503 & 0.139 \\
    \hdashline
        Comet 22 & 0.010 & 0.006 & 0.593 & 0.142 \\
        CometKiwi 22 & 0.007 & 0.003 & 0.415 & 0.071 \\
        CometKiwi 22 XXL & 0.017 & 0.010 & 0.489 & 0.122 \\
    \hdashline
        Gemma 3 (4B) & 0.037 & 0.101 & 0.212 & 0.231 \\
    \hdashline
        BLEU (sent. level) & 1.201 & 0.755 & 0.383 & 0.099 \\
        ChrF (sent. level) & 1.088 & 0.657 & 0.149 & 0.090 \\
        \bottomrule
    \end{tabular}
}
    \caption{Normalized and unnormalized sensitivity against adequacy and fluency.}
    \label{tab:sens_analysis}
\vspace{-10pt}
\end{table}

\subsection{Sensitivity Analysis}
In~\S\ref{sec:pab} and~\S\ref{sec:spa_pane}, we illustrate the position of each metric in the adequacy--fluency tradeoff by their system-level ranking prediction. Here, we aim to analyze how each metric reacts to an adequacy or fluency error.

We do this by controlling one aspect while varying the other. Take adeqacy as an example. Given a source segment, we consider pairs of translation candidates that share the same Fluency~MQM but differ in Adequacy~MQM. For a metric, we report $\frac{\Delta\;\text{metric score}}{\Delta\;\text{Adequacy MQM}}$, averaged over all our translation pairs. This measures the expected change in the metric score per one-point difference in Adequacy~MQM. 

We also report a normalized version of the above measure  by multiplying it with~$\frac{\sum \sigma (\text{Adequacy MQM})}{\sum \sigma (\text{metric score})}$, where $\sigma(\cdot)$ is the standard deviation over different candidate translations given a segment, and the sum is taken over all segments. This scaling enables meaningful comparison across metrics with different output scales. 

Table~\ref{tab:sens_analysis} reports the results for both adequacy and fluency. Here, we can interpret the bias of a metric by  comparing the sensitivity to adequacy and that to fluency. Generally, our findings in previous subsections are also observed in this experiment, except for the normalized sensitivity of FluencyX, which requires further investigation.

\medskip
Overall, this section provides three analysis protocols (PA Breakdown, SPA Plain, and Sensitivity Analysis) to study the position of different evaluation metrics within the adequacy--fluency tradeoff. Our key findings in this section include: (1) Most of the translation metrics are biased toward adequacy, and (2) For the commonly used MetricX and Comet metrics, the former exhibits a more balanced behavior than the latter. Consistent observations in different protocols cross-validate the reliability of our findings.

\section{Conclusion}

In this work, we investigate the adequacy--fluency tradeoff in translation. While this tension is well-documented at the level of translation output~\cite{flamich2025feedbirdsscoreaccuracynaturalness}, we show that it also manifests severely at the levels of evaluation and meta-evaluation.

Through empirical analysis of WMT meta-evaluation protocols, we uncover a systematic bias toward adequacy, driven by the composition of meta-evaluation datasets. We also propose a method that can control the balance between adequacy and fluency.

We further analyze the placement of popular translation evaluation metrics along the adequacy–fluency tradeoff and find that most metrics lean toward adequacy. 

We argue that the adequacy–fluency tradeoff is a critical yet under-recognized matter in translation evaluation and meta-evaluation. While we do not take a stance on how to deal with such bias, our primary contribution is to raise awareness of this matter within the community.

We discuss future work in Appendix~\ref{sec:fw}.

\section*{Limitations}
Our work provides an extensive investigation of adequacy–fluency trade-offs in the evaluation and meta-evaluation of machine translation. However, it also has limitations.

First, our work is based on MQM data, which includes human evaluation that is inherently noisy. It places limits on the generality of the conclusion we draw in this paper. To mitigate this, we report results macro-averaged over five translation datasets. The results on individual dataset are considerably noisier, as shown in  
Appendix~\ref{appendix:pereval}.

Second, this study covers only a limited set of language pairs and systems, owing to the limited availability of MQM data. Expanding the evaluation data would yield more robust conclusion.

Third, we only analyze extrinsic bias (caused by the composition of translation systems in meta-evaluation), while not addressing the intrinsic bias (caused by the design of the MQM framework). Studying the intrinsic bias is highly impactful and warrants further efforts.

\bibliography{anthology,custom}
\bibliographystyle{acl_natbib}

\appendix

\begin{table*}[t]
\centering
\resizebox{\linewidth}{!}{
\begin{tabular}{l|ll|l}
\toprule
\multicolumn{1}{l|}{\textbf{Accuracy errors}} & \multicolumn{2}{l|}{\textbf{Fluency errors}} & \multicolumn{1}{l}{\textbf{Other}} \\ \midrule
Accuracy/Addition & Fluency/Grammar & Style/Archaic or obscure word choice & Other \\
Accuracy/Creative Reinterpretation & Fluency/Inconsistency & Style/Bad sentence structure & Source issue \\
Accuracy/Gender Mismatch & Fluency/Punctuation & Style/Unnatural or awkward & \\
Accuracy/Mistranslation & Fluency/Register & Locale convention/Address format & \\
Accuracy/Omission & Fluency/Spelling & Locale convention/Currency format & \\
Accuracy/Source language fragment & Fluency/Text-Breaking & Locale convention/Time format & \\
Non-translation! & Terminology/Inconsistent & Terminology/Inappropriate for context & \\
\bottomrule
\end{tabular}}
\caption{\citet{flamich2025feedbirdsscoreaccuracynaturalness}'s categorization of MQM errors, used for En--De, Ja--Zh, and Zh--En translation directions.}
\label{tab:mqm_error_classification}
\end{table*}

\begin{table*}[t]
\centering
\resizebox{.68\linewidth}{!}{
\begin{tabular}{l|ll|l}
\toprule
\multicolumn{1}{l|}{\textbf{Accuracy errors}} & \multicolumn{2}{l|}{\textbf{Fluency errors}} & \multicolumn{1}{l}{\textbf{Other}} \\ \midrule
Addition & Capitalization & Date-time format & Other \\
Agreement & Inconsistency & Lacks creativity & Source issue \\
Do not translate & Grammar & Measurement format & \\
Mistranslation & Number format & Punctuation & \\
MT hallucination & Register & Spelling & \\
Omission & Unnatural flow & Whitespace & \\
Untranslated & Word order & Wrong language variety & \\
Wrong named entity &&&\\
Wrong term &&&\\
\bottomrule
\end{tabular}
}
\caption{\citet{flamich2025feedbirdsscoreaccuracynaturalness}'s categorization of MQM errors, used for the En--Es translation direction.}
\label{tab:mqm_error_classification_enes}
\end{table*}

\begin{table*}[t!]
\centering
\resizebox{.8\linewidth}{!}{
\begin{tabular}{l|r|r|r|r|r}
\toprule
Year & 2023 & 2023 & 2024 & 2024 & 2024 \\
Language pair & En--De & Zh--En & En--De & En--Es & Ja--Zh \\
\midrule
Mean All~MQM & 6.17 & 2.51 & 2.50 & 0.58 & 2.92 \\
Mean Adequacy~MQM & 3.97 & 1.62 & 1.38 & 0.45 & 2.55 \\
Mean Fluency~MQM & 1.99 & 0.89 & 1.10 & 0.13 & 0.35 \\
Mean non-zero All~MQM & 10.02 & 5.20 & 4.95 & 3.00 & 6.00 \\
Mean non-zero Adequacy~MQM & 9.39 & 6.21 & 5.63 & 3.91 & 6.38 \\
Mean non-zero Fluency~MQM & 4.23 & 2.36 & 2.82 & 1.37 & 2.07 \\
\midrule
\# segments & 5520 & 1954 & 8766 & 8722 & 7840 \\
\# segments w/o errors (All~MQM = 0) & 1350 & 350 & 3901 & 6672 & 3842 \\
\# segments w/ errors (All~MQM > 0) & 4170 & 1604 & 4865 & 2050 & 3998 \\
\# segments w/ adequacy errors (Adequacy~MQM > 0) & 2919 & 891 & 2738 & 1393 & 3354 \\
\# segments w/ fluency errors (Fluency~MQM > 0) & 3149 & 1250 & 3462 & 849 & 1325 \\
\bottomrule
\end{tabular}}
\caption{Dataset statistics by language pair.}
\label{tab:datasetstatistics}
\end{table*}

\section{Adequacy~MQM and Fluency~MQM Categorization}
\label{appendix:error-classification}

As mentioned in \S\ref{sec:expdataset}, we consider Adequacy~MQM and Fluency~MQM separately, following the categorizations provided in \citet{flamich2025feedbirdsscoreaccuracynaturalness}. Table~\ref{tab:mqm_error_classification} shows the categorization used for En--De, Ja--Zh, and Zh--En. Specially, En--Es follows the categorization in Table~\ref{tab:mqm_error_classification_enes}.

\section{Dataset Statistics}
\label{appendix:dataset_statistics}

Table~\ref{tab:datasetstatistics} summarizes key statistics for each evaluation dataset in our study.

\section{PA vs. SPA}
\label{sec:pavsspa}

In the main paper, we mention that PA and SPA could behave differently, and that it is necessary to consider both PA and SPA in our study. In this appendix, we use a toy example to illustrate the potential disagreement of PA and SPA when assessing whether a metric is biased toward Adequacy~MQM or Fluency~MQM.

Table~\ref{tab:PAvsSPA} shows the details of our example. In this case, PA suggests that the metric is biased toward adequacy, because PA only judges based on the sign. However, SPA suggests the opposite, as it concerns the closeness of the score difference. See \S\ref{sec:SPA} for details.

Therefore, we use both of the metrics, and design dedicated experiments for them. Luckily, our findings are generally consistent under PA and SPA, suggesting that they are reliable meta-metrics in our study.

\section{Measuring Extrinsic Bias without Equal-Variance Assumption}
\label{sec:alteranova}

In \S\ref{sec:measuringimplicit}, for simplicity we assume that all translation systems share the same variance in their Adequacy~MQM scores and the same variance in their Fluency~MQM scores. In this part, we drop this assumption and use \citet{c0eb885e-c393-3777-9bf1-621bcb49b979}'s ANOVA approach (which does not require the equal-variance assumption) to calculate F-statistics and $B$ values.

Table~\ref{tab:measuringimplicitalternova} reports our results. The findings are consistent with those reported in Table~\ref{tab:stds} under the equal-variance assumption. We nevertheless prefer the standard approach presented in the main text due to its greater simplicity and numerical stability. Note that we observe undefined $p$-values in our preliminary experiments for some synthetic datasets in \S\ref{sec:reducingimplicit}, if we do not have the assumption.

\section{Analysis Results for Setup~7}
\label{appendix:setup7}

\begin{figure*}[t!]
\begin{center}
\includegraphics[width=.88\linewidth]{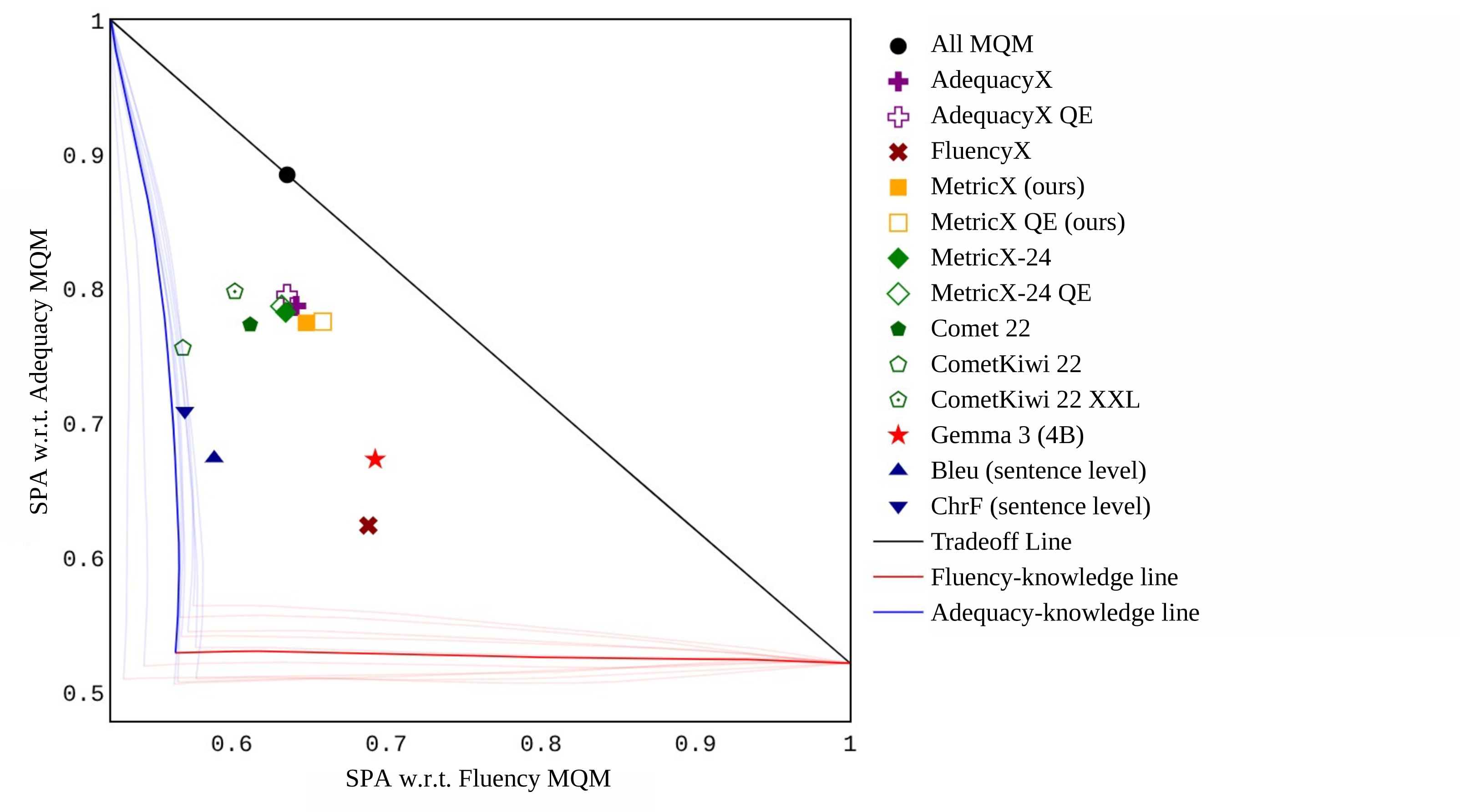}
\end{center}
\vspace{-10pt}
\caption{SPA plane using Setup~7, macro-averaged across five evaluation sets. Each shadow line represents the combination of Adequacy~MQM or Fluency~MQM with a specific random noise instance. The solid red and blue lines are the average of the shadow lines.}
\label{fig:SPA7}
\end{figure*}

In this section, we present the results of PA Breakdown (\S\ref{sec:pab}) and SPA Plane (\S\ref{sec:spa_pane}) based on Setup~7 in Table~\ref{tab:synthesized_systems}. We include this setup for the sake of completeness, as it closely matches Row~6 (used in our main analysis) in terms of $B$ values. Table~\ref{tab:Concordance7}, Table~\ref{fig:PAB7}, and Figure~\ref{fig:SPA7} present the concordance/discordance ratio, PA Breakdown, and SPA Plane results, respectively, for Setup~7. All our findings based on Setup~6 also hold in this setup.

\section{Adequacy--Fluency Tradeoff Results on Individual Evaluation Sets}
\label{appendix:pereval}

In \S\ref{sec:exptradeoffeval}, we report results macro-averaged across five evaluation sets (of different language pairs). In this part, we present the results on each evaluation set.

\begin{figure*}[t]
\begin{center}
\includegraphics[width=\linewidth]{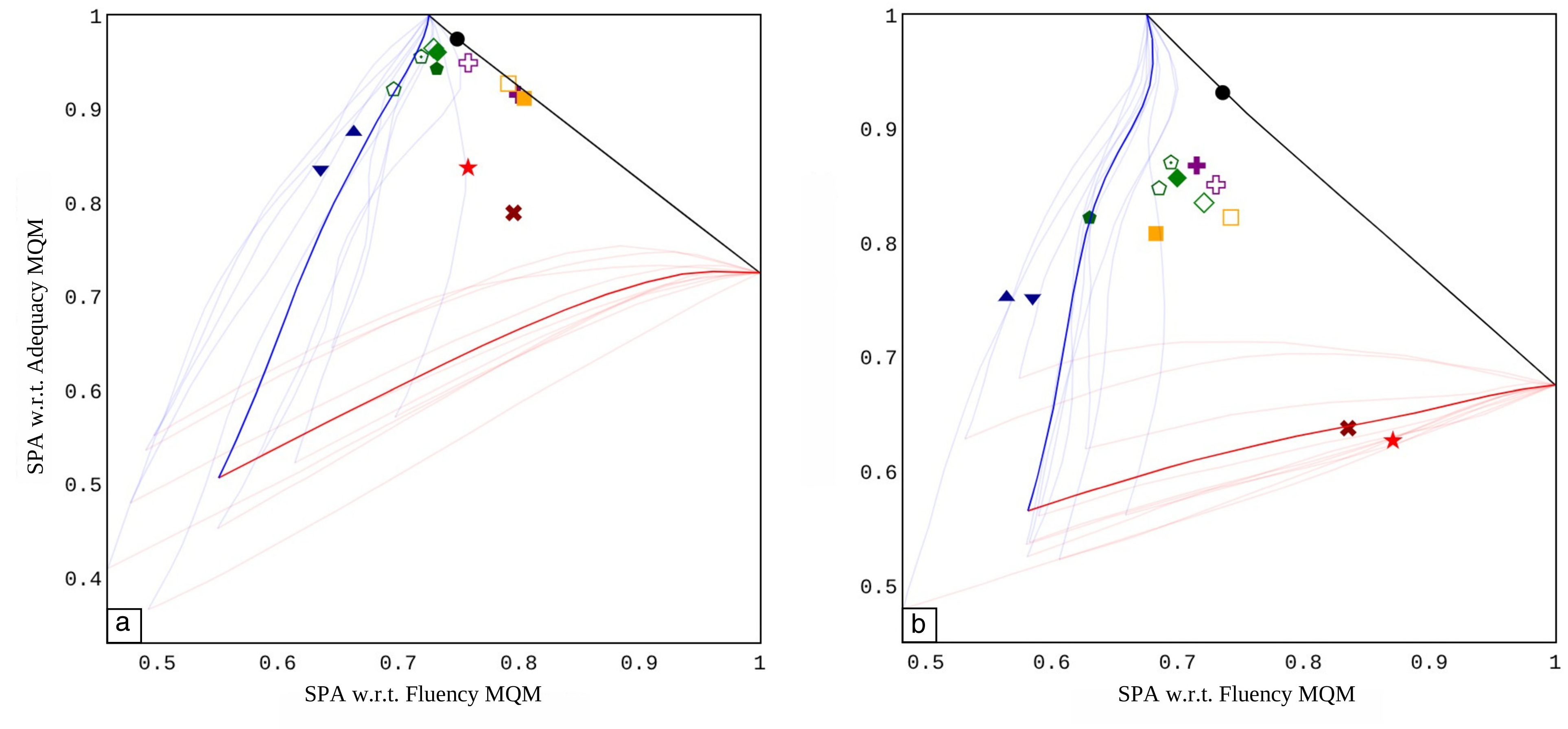}
\includegraphics[width=\linewidth]{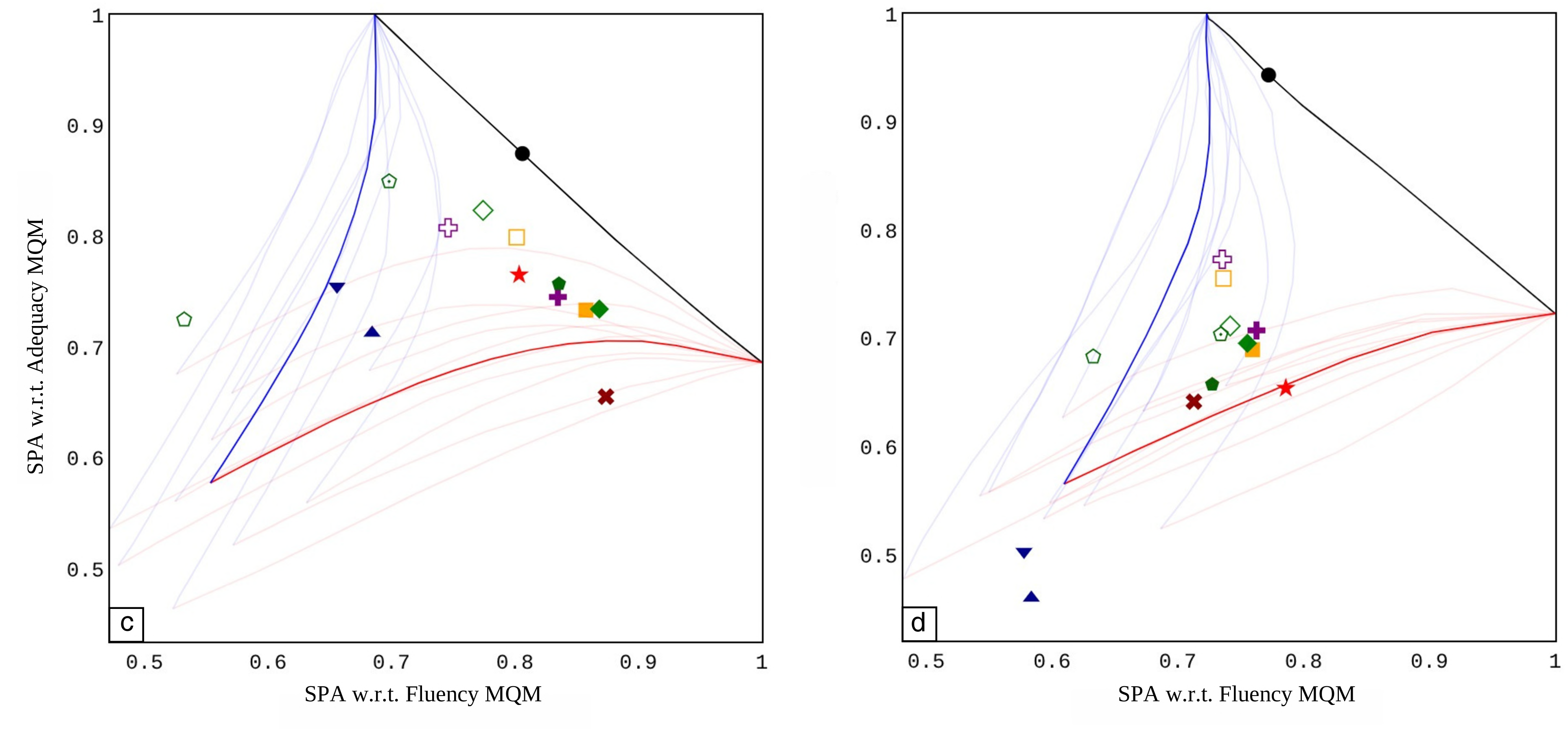}
\includegraphics[width=\linewidth]{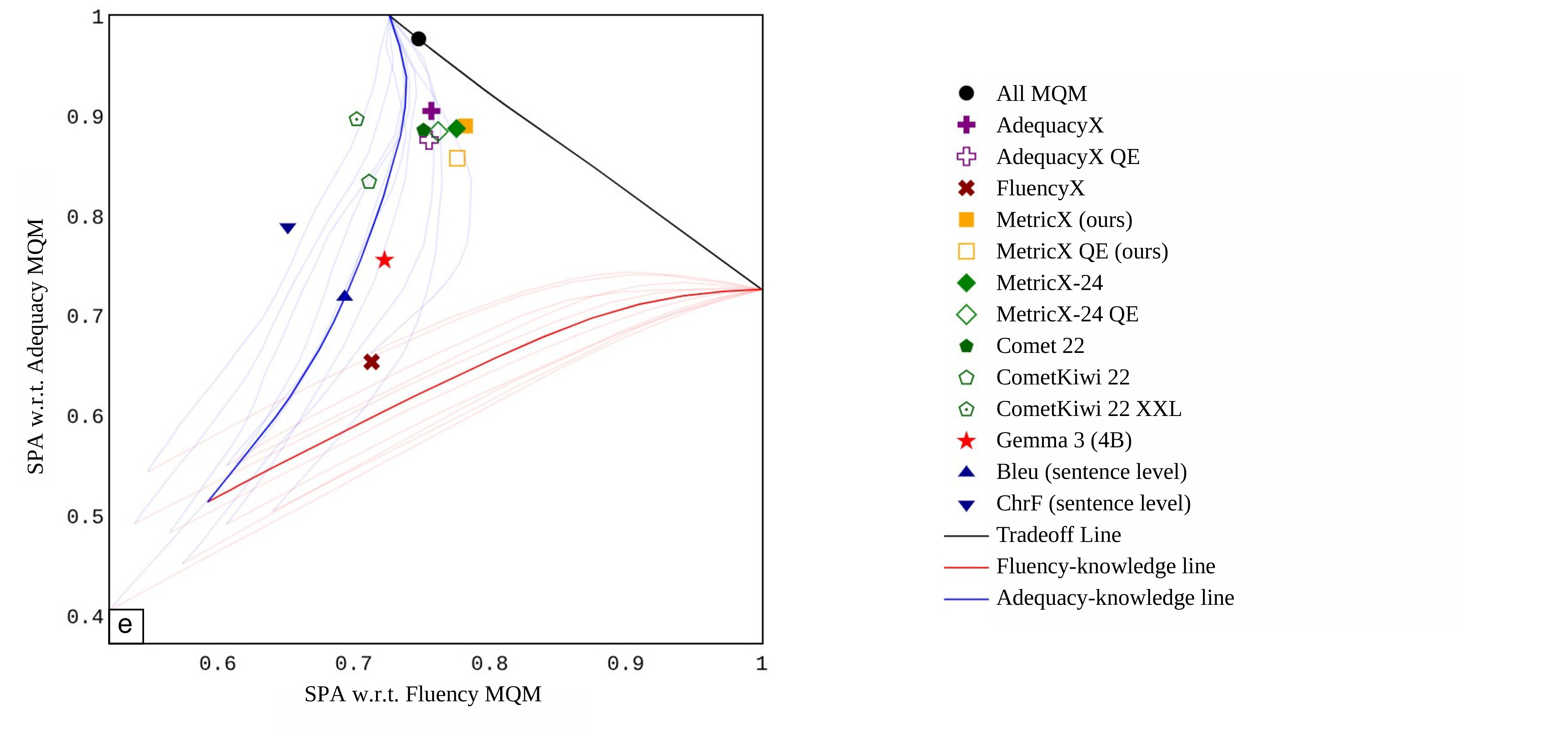}
\end{center}
\caption{SPA plane based on the original WMT setup for (a) En--De 2023, (b) Zh--En 2023, (c) En--De 2024, (d) En--Es 2024, and (e) Ja--Zh 2024. The legend is shared for all the figures.}
\label{fig:SPAplanepereval}
\end{figure*}

In Table~\ref{fig:PABpereval} and Figure~\ref{fig:SPAplanepereval}, we report PA-Breakdown and SPA Planes results based on the original WMT setup, presented separately for each evaluation set. The results exhibit diverse behaviors across evaluation sets. Thus, we perform macro-average in our main paper. The noisy phenomenon requires further investigation.

\section{Future Work}
\label{sec:fw}

Our study opens several avenues for future research, discussed below.

First, we reveal the potential imbalance in the translation meta-evaluation and highlight the importance of understanding this phenomenon~(\S\ref{sec:imbalance}). We propose a statistical measure to quantify this imbalance and design a method to control it through data synthesis. However, there is room for improvement in both measuring the imbalance and debiasing the meta-evaluation. In particular, we are interested in exploring theoretical approaches to debiasing the meta-evaluation outputs by normalizing scores in a post hoc manner, without altering the datasets.

Second, we develop dedicated translation metrics, AdequacyX and FluencyX, to independently assess adequacy and fluency~(\S\ref{sec:underanalmetrics}). We aim to further improve this separation, as it facilitates adequacy--fluency analyses, especially in scenarios where MQM annotations are not available.

Third, as we are now in the era of large language models (LLMs), it is interesting to study the LLM-as-a-judge  for translation through the lens of the adequacy--fluency tradeoff, for example, understanding and steering the bias of LLM-as-a-judge.

\begin{table*}[t!]
    \centering
    \begin{minipage}[!t]{0.47\linewidth}
        \begin{tabular}{l|c|c|c}
            \toprule
             & Sys 1 & Sys 2 & $\Delta$ \\
            \midrule
            Adequacy~MQM & $8.0$ & $6.0$ & $+2.0$\tikzmark{a}\\
            Fluency~MQM & $6.0$ & $6.5$ & $-0.5$\tikzmark{b}\\
            \hline
            Metric & $6.3$ & $6.0$ & $+0.3$\tikzmark{c}\\
        \bottomrule
        \end{tabular}
        \begin{tikzpicture}[remember picture, overlay]
        \draw [<->,black,line width=.75pt] ([xshift=1ex, yshift=1.0ex]pic cs:c) [bend right=55] to node[sloped, yshift=-1ex] {\footnotesize \quad closer} ([xshift=1ex, yshift=.5ex]pic cs:b);
        \draw [<->,black,line width=.75pt] ([xshift=3ex, yshift=-.3ex]pic cs:c) [bend right=90] to node[sloped, yshift=-1ex] {\footnotesize same sign} ([xshift=3ex, yshift=.5ex]pic cs:a);
        \end{tikzpicture}
        \caption{A toy example illustrating potential disagreement between PA and SPA: PA prefers keeping the same sign, whereas SPA prefers the closer one. Here, all numbers are hypothetical for illustration purposes.}
        \label{tab:PAvsSPA}
    \end{minipage}
    \hfill
    \begin{minipage}[!t]{0.47\textwidth}
        \centering
        \begin{tabular}{l|r|r | r}
        \toprule
              & \multicolumn{2}{c|}{F-statistic} & $B(\Delta p)$ \\
             & \multicolumn{1}{c|}{Adequacy} & \multicolumn{1}{c|}{Fluency} & \\
            \midrule
            En--De'23 & $28.9$ & $8.4$ & $0.07^A$ \\
            Zh--En'23 & $84.1$ & $11.9$ & $0.04^A$\\
            En--De'24 & $13.0$ & $9.0$ & $0.04^A$ \\
            En--Es'24 & $26.5$ & $4.2$ & $0.14^A$ \\
            Ja--Zh'24 & $28.8$ & $6.9$ & $0.08^A$ \\
            \hline
            Macro-Avg & $36.3$ & $8.1$ & \\
            \bottomrule
        \end{tabular}
        \caption{\citet{c0eb885e-c393-3777-9bf1-621bcb49b979}'s F-statistics for Adequacy~MQM and Fluency~MQM, and the respective $B$-values, in the original meta-evaluation setup.}
        \label{tab:measuringimplicitalternova}
    \end{minipage}
\end{table*}

\begin{table*}[t]
    \centering
    \begin{minipage}[!t]{0.36\linewidth}
        \centering
        \resizebox{\linewidth}{!}{
        \begin{tabular}{l|c|c}
            \toprule
            Evaluation Set & Concordance & Discordance \\
            \midrule
            En--De'23 & 356 (57\%) & 274 (43\%) \\
            Zh--En'23 & 370 (37\%) & 620 (63\%) \\
            En--De'24 & 692 (54\%) & 583 (46\%) \\
            En--Es'24 & 375 (51\%) & 366 (49\%) \\
            Ja--Zh'24 & 399 (54\%) & 342 (46\%) \\
            \bottomrule
        \end{tabular}}
        \caption{Concordance and discordance between Adequacy~MQM and Fluency~MQM in Setup~7, reported as system pair counts and percentages.}
        \label{tab:Concordance7}
    \end{minipage}
    \hfill
    \begin{minipage}[!t]{0.62\textwidth}
        \centering
        \includegraphics[width=\linewidth]{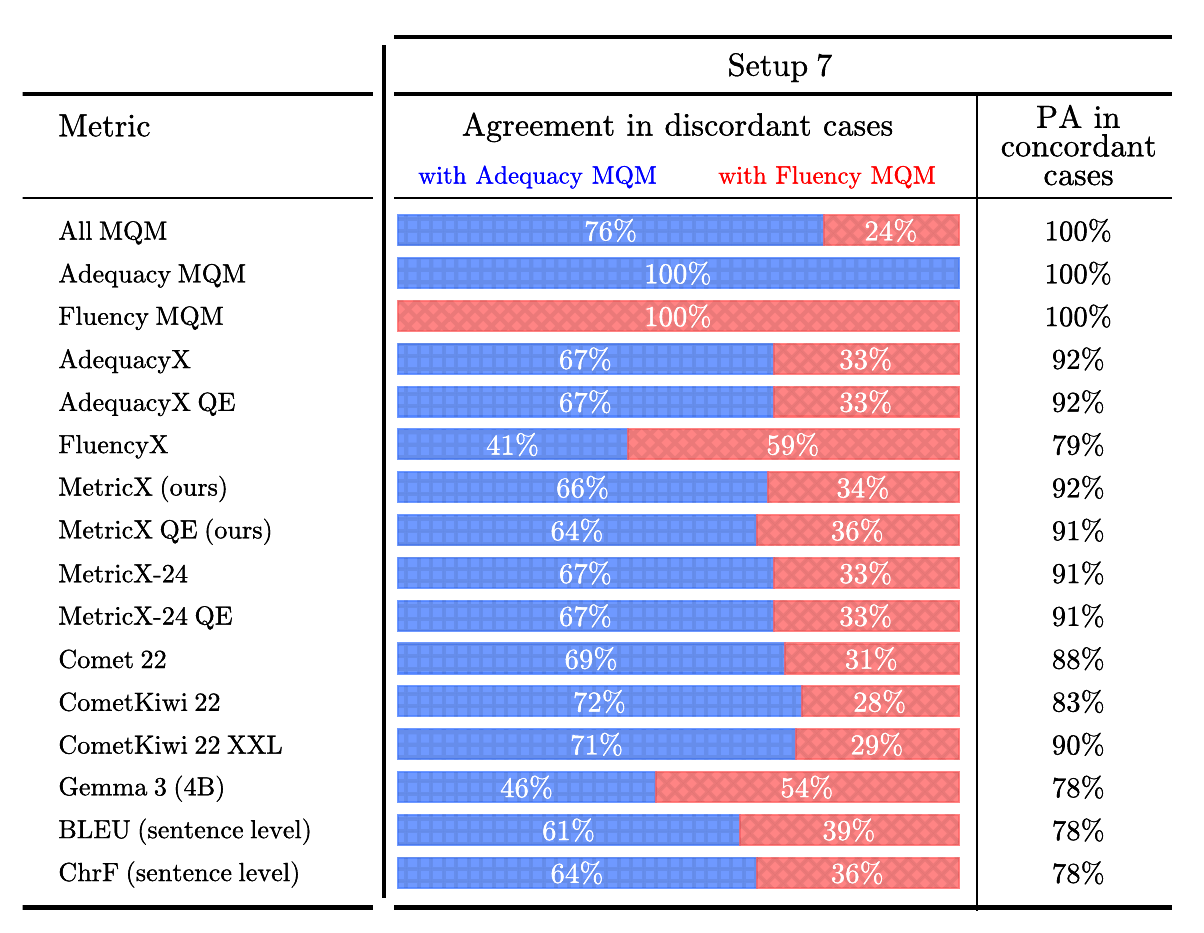}
        \vspace{-20pt}
        \caption{PA breakdown, macro-averaged across five evaluation sets, for Setup~7.}
        \label{fig:PAB7}
    \end{minipage}
\end{table*}

\newpage

\begin{table*}[t]
\begin{center}
\includegraphics[width=\linewidth]{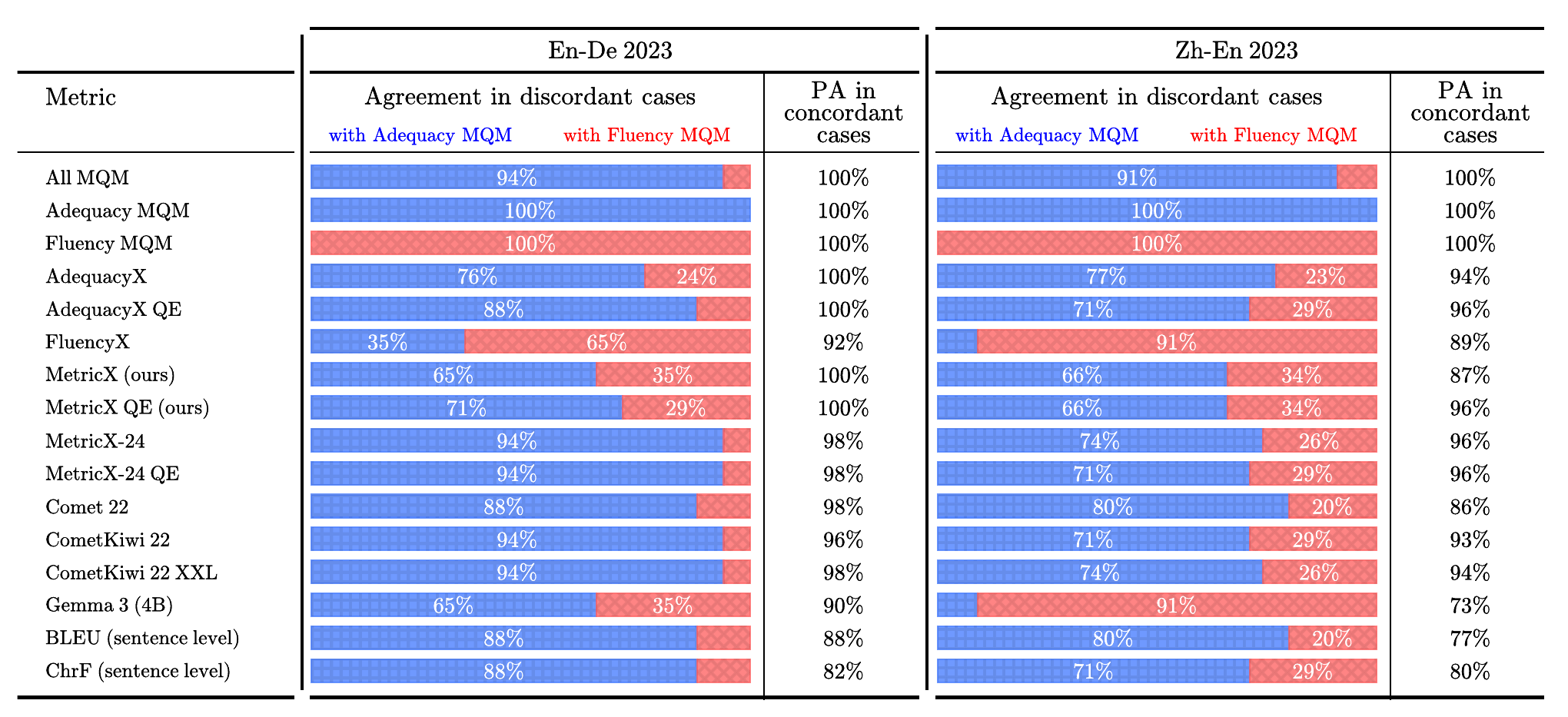}
\includegraphics[width=\linewidth]{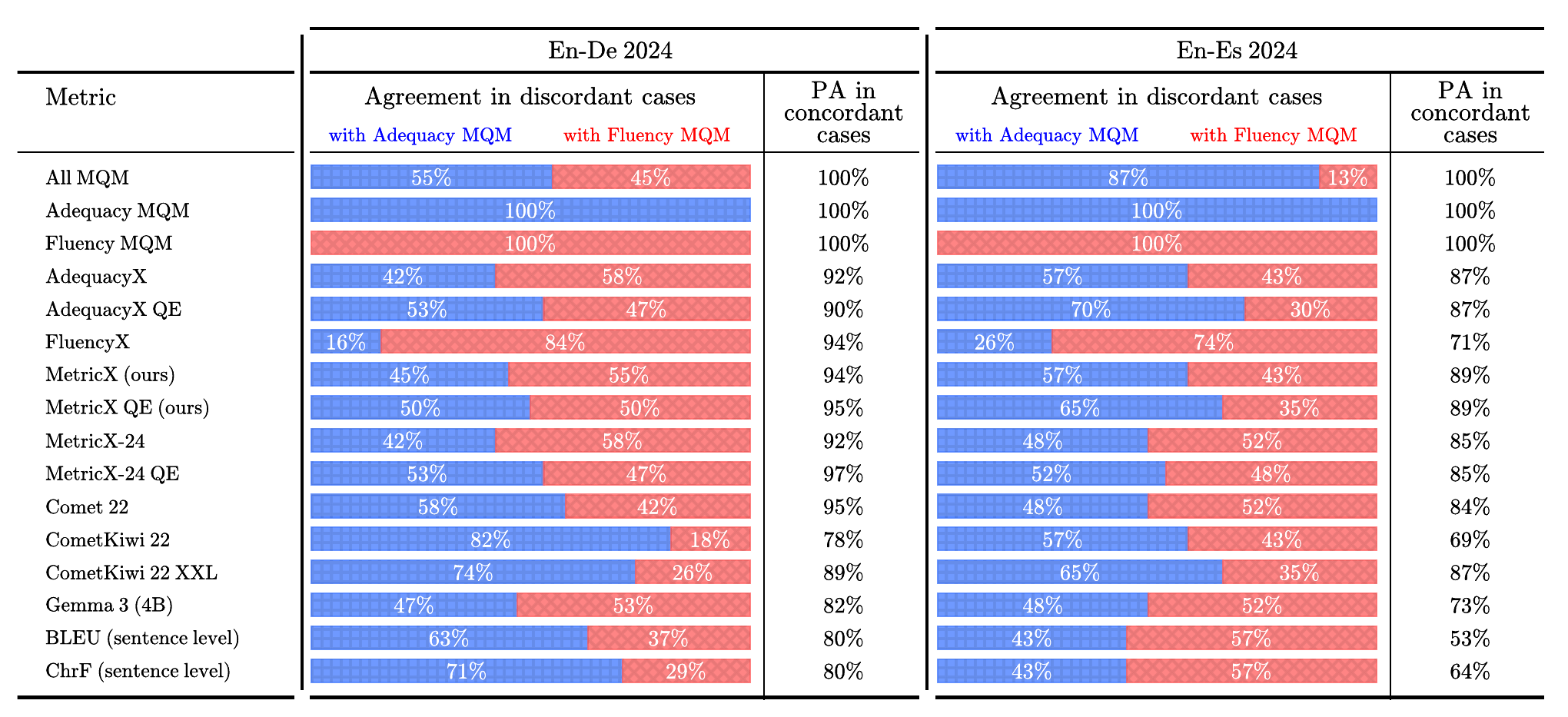}
\includegraphics[width=\linewidth]{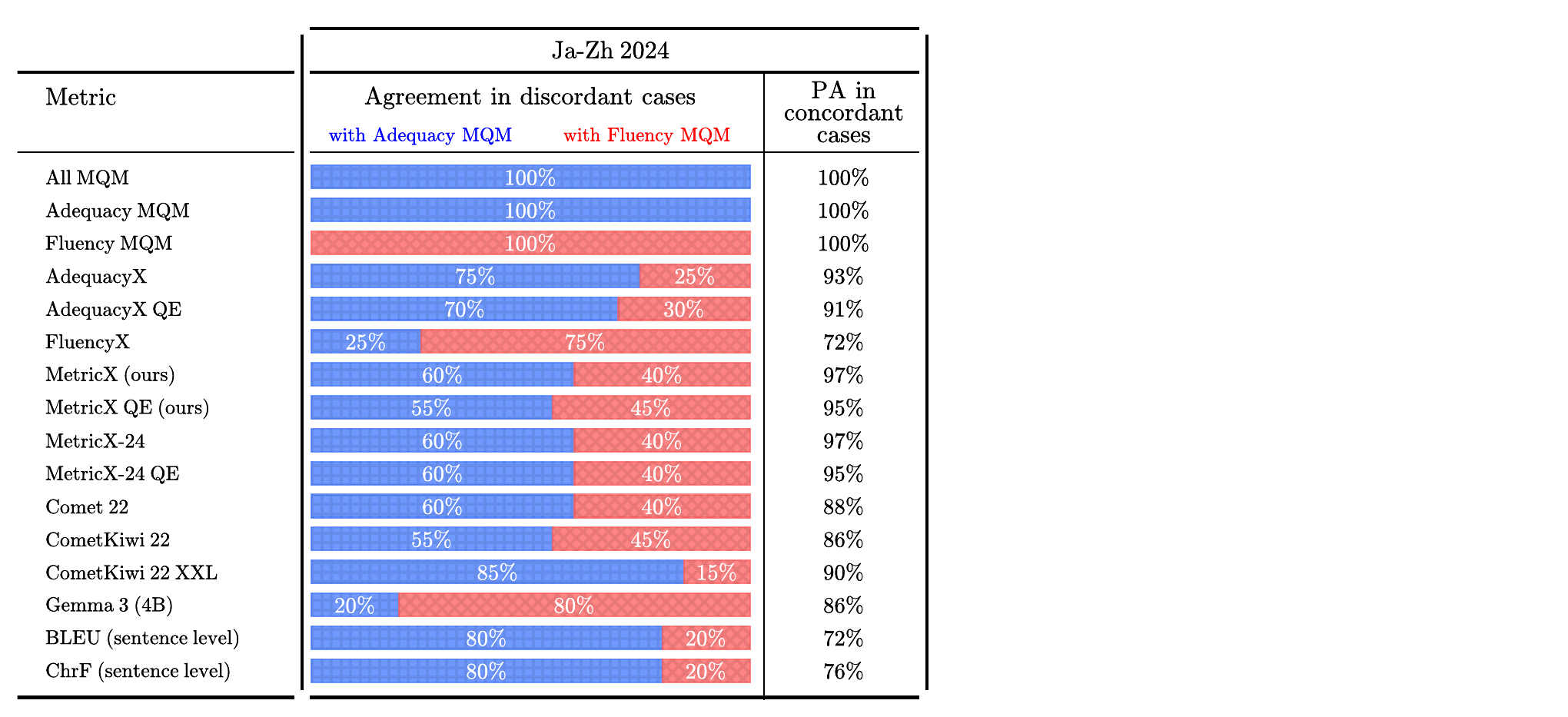}
\end{center}
\caption{PA breakdown, per evaluation set, based on the original WMT setup.}
\label{fig:PABpereval}
\end{table*}

\end{document}